\documentclass[11pt,twocolumn,letterpaper]{article}

\usepackage[utf8]{inputenc}
\usepackage{amsmath,amssymb,amsthm}
\usepackage{graphicx}
\usepackage{booktabs}
\usepackage{multirow}
\usepackage{algorithm}
\usepackage{algpseudocode}
\usepackage{hyperref}
\usepackage{caption}
\usepackage{subcaption}
\usepackage[margin=0.75in]{geometry}
\usepackage{balance}  
\usepackage{tikz}     
\usetikzlibrary{shapes,arrows,positioning,fit,backgrounds,calc}

\definecolor{inputcolor}{RGB}{200,220,255}
\definecolor{gnncolor}{RGB}{255,200,200}
\definecolor{bertcolor}{RGB}{255,200,150}
\definecolor{attncolor}{RGB}{255,220,150}
\definecolor{poolcolor}{RGB}{200,255,200}
\definecolor{outputcolor}{RGB}{220,200,255}

\newcommand{\strongemph}[1]{\textbf{#1}}

\title{Tree Matching Networks for Natural Language Inference:\\
Parameter-Efficient Semantic Understanding via\\
Dependency Parse Trees}

\author{Jason Lunder\\
Eastern Washington University\\
\texttt{jlunder@ewu.edu}}

\date{\today}

\begin{document}

\maketitle

\begin{abstract}
Transformer-based models like BERT achieve high accuracy on Natural Language Inference (NLI) but require hundreds of millions of parameters and extensive pretraining. We investigate whether explicit dependency tree structures can improve parameter efficiency by providing syntactic inductive bias that transformers must learn from scratch.

We adapt Graph Matching Networks to operate on dependency parse trees, creating Tree Matching Networks (TMN), and compare against BERT-based baselines with matched parameters on SNLI and SemEval tasks. TMN substantially outperforms BERT at small scales with reduced memory and training requirements, validating that dependency structure provides measurable benefits.

However, we identify a scaling plateau: increasing TMN parameters 2$\times$ yields minimal improvement, indicating that simple aggregation methods create an architectural bottleneck. These findings motivate hybrid approaches that preserve structural benefits while addressing scalability limitations, such as using graph neural networks to pre-encode dependency trees before transformer-based aggregation with tree-aware positional encodings.

\end{abstract}

\section*{Introduction}

Natural language understanding requires capturing both semantic content and syntactic structure. Modern transformer-based models like BERT \cite{devlin2019bert} achieve impressive results on natural language inference (NLI) tasks but require hundreds of millions of parameters and extensive pretraining. We investigate whether explicit structural representations can improve parameter efficiency while maintaining competitive performance.

Dependency parse trees explicitly encode syntactic relationships as graph edges, potentially providing inductive biases that reduce learning complexity. Consider the sentence ``A dog runs in the park.'' A dependency parser identifies ``dog'' as the nominal subject (nsubj) of ``runs,'' ``in'' as a prepositional modifier, and ``park'' as the object of the preposition. These relationships are encoded as labeled edges in the dependency tree. In contrast, transformer models must learn these relationships implicitly from token co-occurrence patterns across massive corpora.

\subsection*{Research Question}

\strongemph{Can graph neural networks operating on dependency parse trees provide more parameter-efficient sentence embeddings than transformer-based models for natural language inference?}

Specifically, we investigate whether Tree Matching Networks (TMN), our adaptation of Graph Matching Networks \cite{li2019graph} to dependency trees, can outperform BERT at equivalent parameter counts on natural language inference.

\subsection*{Our Approach}

We adapt Graph Matching Networks (GMN) \cite{li2019graph} to operate on linguistic dependency trees with rich node and edge features. We compare two architectural paradigms: matching models that use cross-graph attention between sentence pairs (TreeMatchingNet vs BertMatchingNet), and embedding models that process inputs independently with comparison only at the embedding level (TreeEmbeddingNet vs BertEmbeddingNet). Both TMN and BERT models use similar parameter counts ($\sim$36M for TMN, $\sim$41M for BERT) and undergo identical three-phase training: (1) contrastive pretraining on WikiQS and AmazonQA, (2) task-specific multi-objective InfoNCE training, and (3) direct classification fine-tuning.

\subsection*{Key Findings}

TreeMatchingNet achieves 75.20\% SNLI test accuracy compared to BertMatchingNet's 35.38\%, demonstrating 2.1$\times$ superior performance with comparable parameter counts. This translates to improved parameter efficiency: TMN achieves 2.4$\times$ better efficiency (2.09\% vs 0.86\% accuracy per million parameters). However, we observe a scaling plateau: increasing TMN parameters from 18.8M to 36M (2$\times$) yields minimal improvement (0.2 percentage points), revealing an architectural bottleneck in the aggregation mechanism. Meanwhile, BertMatchingNet exhibits a failure mode where it predicts every test example as entailment, suggesting incompatibility between the cross-attention architecture and BERT's design.

\subsection*{Contributions}

We present an adaptation of Graph Matching Networks to linguistic dependency trees with rich features combining BERT embeddings, POS tags, and morphological annotations for natural language inference. Our evaluation compares tree-based GNN and BERT architectures at matched parameter counts ($\sim$36M) with identical training protocols, enabling controlled analysis of structural inductive bias. We identify a scaling plateau where 2$\times$ parameter increases yield minimal improvement, suggesting the aggregation mechanism as an architectural bottleneck. Through systematic comparison of matching versus embedding architectures, we isolate the contribution of cross-graph attention and observe compatibility differences between tree-based and transformer-based approaches.

\section*{Related Work}

\subsection*{Natural Language Inference}

The Stanford Natural Language Inference (SNLI) corpus \cite{bowman2015large} contains 570K sentence pairs labeled as entailment, contradiction, or neutral. This dataset has become a standard benchmark for evaluating semantic understanding in NLP systems. State-of-the-art models like EFL with RoBERTa-large \cite{wang2021entailment} achieve approximately 90\% accuracy but require 355M parameters. Our goal is to achieve competitive accuracy with substantially fewer parameters (10$\times$ reduction) by leveraging structural inductive biases rather than relying solely on model scale and extensive pretraining.

\subsection*{Graph Neural Networks and Graph Matching}

Graph neural networks (GNNs) have emerged as effective models for learning on structured data \cite{scarselli2009graph}. Modern GNN architectures \cite{li2015gated,kipf2016semi,velickovic2017graph} compute node representations through iterative message passing along graph edges, aggregating information from local neighborhoods. These models are permutation-invariant by design and have been successfully applied to molecular property prediction, knowledge graph reasoning, and program analysis.

\strongemph{Graph Matching Networks} \cite{li2019graph} introduced a cross-graph attention mechanism for computing similarity between pairs of graphs. The GMN architecture consists of three key components: (1) graph propagation layers that update node representations via message passing, (2) cross-graph attention that computes attention-weighted matchings between nodes in different graphs, and (3) aggregation that produces graph-level embeddings. GMNs have been applied to graph edit distance learning, molecular similarity search, and binary function similarity analysis.

\strongemph{Our extension:} We adapt GMN principles to linguistic dependency trees, incorporating domain-specific features (BERT embeddings, POS tags, morphology) and developing a multi-objective training protocol for multi-class NLI.

\subsection*{Dependency Parsing for NLU}

Dependency trees have been used in NLP for semantic role labeling \cite{gildea2002automatic}, relation extraction \cite{bunescu2005shortest}, and machine translation \cite{ding2005machine}. Early neural approaches combined dependency structures with neural models: Recursive neural networks \cite{socher2013recursive} and Tree-LSTMs \cite{tai2015improved} process parse trees in a bottom-up fashion, demonstrating that syntactic structure improves semantic representations. However, these approaches struggled with sequential processing bottlenecks and limited scalability.

\subsection*{Hybrid Structure-Transformer Approaches}

Recent work explores combining dependency structures with transformers to preserve syntactic inductive bias while achieving scalability. Syntax-BERT \cite{sachan2021syntax} proposes late fusion (applying graph neural networks to transformer outputs) and joint fusion (interleaving GNN layers within transformer blocks). Dependency Transformer Grammars \cite{zhao2024dependency} modify transformer attention masks to simulate dependency transition systems, encoding syntactic constraints directly in attention patterns. Stack Attention \cite{haviv2024transformer} incorporates implicit syntactic structure through stack operations in attention mechanisms, learning syntax without explicit parse trees.

These approaches differ from our work in architectural positioning: they modify transformer attention mechanisms or post-process transformer outputs, while our proposed approach uses graph neural networks as a preprocessing step to create structurally-enriched node embeddings that serve as input to standard transformer aggregation. The current paper validates that dependency tree structure provides measurable benefits worth preserving in such hybrid architectures.

\subsection*{Parameter-Efficient NLP}

Growing interest in parameter-efficient NLP has led to various approaches: knowledge distillation \cite{sanh2019distilbert} transfers knowledge from large models to smaller ones, pruning \cite{michel2019sixteen} removes unnecessary parameters, and efficient architectures like ALBERT \cite{lan2020albert} reduce parameters through factorization and sharing. Our approach is complementary: we explore whether structural inductive biases can reduce parameter requirements from the ground up, rather than compressing existing large models.

\subsection*{Distance Metric Learning}

Metric learning aims to learn distance functions that group similar examples together and separate dissimilar ones. Siamese networks \cite{bromley1994signature} apply the same network to two inputs independently and compute similarity from the resulting embeddings. These architectures have achieved success in face verification \cite{chopra2005learning,sun2014deep} and image matching \cite{zagoruyko2015learning}.

Our graph matching models differ from standard Siamese networks: rather than processing inputs independently and comparing only at the output, we employ cross-graph attention throughout the propagation process. This enables early information fusion and allows the model to adjust representations based on what they are being compared to.

\section*{Tree-Based Similarity Learning for Natural Language Inference}

\subsection*{Problem Formulation}

Natural Language Inference (NLI) involves determining the relationship between a premise sentence $P$ and hypothesis sentence $H$, with labels in \{Entailment ($+1$), Neutral ($0$), Contradiction ($-1$)\}. We use the Stanford Natural Language Inference (SNLI) corpus \cite{bowman2015large} as our primary evaluation benchmark. SNLI is a standard dataset for evaluating semantic understanding in NLP systems, containing sentence pairs derived from image captions with crowd-sourced annotations. Performance on SNLI demonstrates a model's ability to understand entailment relationships, which requires capturing both semantic content and logical reasoning.

For generalization evaluation, we also use the SemEval Semantic Textual Similarity benchmark, which frames the task as a 2-class similarity problem (similar vs dissimilar). This tests whether structural benefits transfer beyond entailment to broader semantic similarity judgments.

\begin{table*}[htbp]
\centering
\caption{Dataset statistics: original corpus sizes, converted tree counts, and training usage. Contrastive pretraining and SNLI training use sampled batches (600/epoch); SemEval uses the complete dataset.}
\label{tab:dataset_stats}
\footnotesize
\begin{tabular}{lcccccc}
\toprule
\multirow{2}{*}{Dataset} & \multirow{2}{*}{Task} & \multicolumn{2}{c}{Original Corpus} & \multicolumn{2}{c}{Converted to Trees} & \multirow{2}{*}{Training Usage} \\
\cmidrule(lr){3-4} \cmidrule(lr){5-6}
 & & Train & Dev/Test & Train & Dev/Test & \\
\midrule
\multicolumn{7}{l}{\textit{Contrastive Pretraining (Large-Scale Corpus)}} \\
WikiQS & Question Sim. & $\sim$28.8M & -- & 1,850,142 & 261,292 / 575,121 & \multirow{2}{*}{\shortstack[l]{600 batches/epoch\\(153,600 samples)}} \\
AmazonQA & Q\&A Pairs & $\sim$923K & -- & 669,943 & 83,138 / 85,089 & \\
\cmidrule{6-6}
\multicolumn{5}{r}{\textit{Combined Contrastive Train:}} & \textbf{2,520,085} & \\
\midrule
\multicolumn{7}{l}{\textit{Primary Training and Fine-Tuning (Task-Specific)}} \\
SNLI & Entailment & 550,152 & 10K / 10K & 540,803 & 9,969 / 9,960 & \shortstack[l]{600 batches/epoch\\(153,600 samples)} \\
SemEval & Similarity & 3,000 & 750 / 6,750 & 2,967 & 750 / 6,638 & Complete dataset \\
\bottomrule
\end{tabular}
\end{table*}

\noindent \textbf{Data Preparation.} Raw text datasets were converted to dependency tree representations using TMN\_DataGen (Section 3.1.1). For large-scale contrastive pretraining (WikiQS + AmazonQA), computational constraints limited full-dataset conversion and training: we converted 2.52M training pairs and randomly sample 600 batches per epoch (153,600 pairs, batch size 256) without replacement within each epoch. This sampling strategy processes $\sim$6\% of the contrastive corpus per epoch. Each epoch uses a different random sample from the full pool. For SNLI, we apply the same sampling strategy (600 batches per epoch) to enable comparable training across all phases. For SemEval, we use the complete dataset without sampling due to its smaller size (2,967 training pairs).

\subsection*{Data Processing Pipeline}

We developed a custom data processing pipeline (TMN\_DataGen) that transforms raw text into dependency trees with rich linguistic features.

\subsubsection*{Preprocessing}

Text normalization includes Unicode handling, case preservation, and word boundary detection. We employ configurable strictness levels (0-3) to handle varying amounts of noise in different datasets.

\subsubsection*{Dependency Parsing}

We use a multi-parser approach to combine the strengths of different parsing systems. DiaParser (based on Electra-base) provides accurate dependency tree structures and dependency relation labels, which form the graph edges in our model. SpaCy provides complementary linguistic annotations including part-of-speech tags, lemmas, and morphological features. This multi-parser strategy leverages DiaParser's superior accuracy for syntactic structure while benefiting from SpaCy's comprehensive linguistic feature extraction.

\subsubsection*{Feature Extraction}

Each dependency tree node is represented by an 804-dimensional feature vector combining semantic and syntactic information. BERT-base-uncased provides 768-dimensional contextual word embeddings that capture semantic content. Part-of-speech tags contribute 17 dimensions (one-hot encoded categories including NOUN, VERB, ADJ, ADV, etc.) extracted from SpaCy. Morphological features contribute 19 dimensions (one-hot encoded attributes such as Number, Tense, Person, Mood, etc.) also from SpaCy. Dependency edges are represented by 70-dimensional one-hot vectors encoding the dependency relation type (nsubj, dobj, amod, etc.) provided by DiaParser.

This combination provides both semantic representations (BERT embeddings) and explicit syntactic annotations (POS tags, morphology, dependency relations), allowing the model to leverage both learned distributional semantics and linguistic structure.


\subsection*{Model Architectures}

We compare two architectural paradigms - \strongemph{matching} (with cross-graph attention) and \strongemph{embedding} (independent processing) - for both tree-based and transformer-based models.

\subsubsection*{Tree Matching Network (TreeMatchingNet)}

The Tree Matching Network adapts Graph Matching Networks to dependency trees. The architecture consists of four main components:

\strongemph{1. TreeEncoder:} Maps input features to initial node and edge representations:
\begin{align}
h_i^{(0)} &= \text{MLP}_{\text{node}}(x_i), \quad \forall i \in V \\
e_{ij} &= \text{MLP}_{\text{edge}}(x_{ij}), \quad \forall (i,j) \in E
\end{align}

The node encoder $\text{MLP}_{\text{node}}$ transforms input node features to hidden node states. The edge encoder $\text{MLP}_{\text{edge}}$ transforms edge features to hidden edge states.

\strongemph{2. Graph Propagation:} We employ $T$ propagation layers. Following the GMN architecture \cite{li2019graph}, we use \emph{shared} propagation parameters across all $T$ layers (contrast with \emph{unshared} variants where each layer has independent parameters). Each layer performs message passing along tree edges followed by cross-graph attention:

\begin{align}
m_{j \rightarrow i} &= f_{\text{message}}(h_i^{(t)}, h_j^{(t)}, e_{ij}) \\
\mu_{j \rightarrow i} &= f_{\text{match}}(h_i^{(t)}, h_j^{(t)}) \quad \text{[}j \text{ from other graph]} \\
h_i^{(t+1)} &= f_{\text{node}}\left(h_i^{(t)}, \sum_{j} m_{j \rightarrow i}, \sum_{j'} \mu_{j' \rightarrow i}\right)
\end{align}

Here, $f_{\text{message}}$ is an MLP that concatenates source node, target node, and edge features. The function $f_{\text{match}}$ implements cross-graph attention (detailed below). The function $f_{\text{node}}$ is a GRU cell that updates node states based on both within-graph messages and cross-graph matching signals.

\strongemph{3. Cross-Graph Attention:} The cross-graph attention mechanism computes how well each node in one graph matches nodes in the other graph:

\begin{align}
a_{j \rightarrow i} &= \frac{\exp(\text{sim}(h_i^{(t)}, h_j^{(t)}))}{\sum_{j'} \exp(\text{sim}(h_i^{(t)}, h_{j'}^{(t)}))} \\
\mu_{j \rightarrow i} &= a_{j \rightarrow i} \cdot (h_i^{(t)} - h_j^{(t)}) \\
\sum_j \mu_{j \rightarrow i} &= h_i^{(t)} - \sum_j a_{j \rightarrow i} h_j^{(t)}
\end{align}

This formulation has a useful property: when two graphs match perfectly and attention weights concentrate on the correct matches, $\sum_j \mu_{j \rightarrow i} \rightarrow 0$, causing the cross-graph communication to vanish. Conversely, differences between graphs are captured in the matching vectors and amplified through propagation.

\strongemph{4. Graph Aggregation:} After propagation, we aggregate node representations to produce graph-level embeddings:

\begin{equation}
h_{\text{graph}} = \text{MLP}_G\left(\sum_{i \in V} \sigma(\text{MLP}_{\text{gate}}(h_i^{(T)})) \odot \text{MLP}(h_i^{(T)})\right)
\end{equation}

The gated weighted sum allows the model to learn which nodes are most important for the graph-level representation. The final $\text{MLP}_G$ projects to the graph representation dimension.

\strongemph{Configuration:} Our specific TreeMatchingNet instantiation uses the hyperparameters shown in Table \ref{tab:tmn_config}. This configuration yields approximately 36M parameters total.

\begin{table}[htbp]
\centering
\caption{TreeMatchingNet configuration.}
\label{tab:tmn_config}
\footnotesize
\begin{tabular}{lc}
\toprule
Component & Value \\
\midrule
Node features & 804 \\
Edge features & 70 \\
Node state dim & 1536 \\
Edge state dim & 768 \\
Prop. layers ($T$) & 5 (shared) \\
Graph rep. dim & 2048 \\
Total params & $\sim$36M \\
\bottomrule
\end{tabular}
\end{table}

\begin{algorithm}[t]
\caption{Tree Matching Network Forward Pass}
\label{alg:tmn}
\begin{algorithmic}[1]
\Require Trees $T_A$ and $T_B$ with node and edge features
\Ensure Graph embeddings $e_A, e_B \in \mathbb{R}^d$ where $d$ is graph representation dimension
\State $N_A, E_A \gets \text{TreeEncoder}(T_A)$
\State $N_B, E_B \gets \text{TreeEncoder}(T_B)$
\For{layer $l = 1$ to $T$}
    \State $N_A, E_A \gets \text{Propagation}_l(N_A, E_A)$ \Comment{Within-graph messages}
    \State $N_B, E_B \gets \text{Propagation}_l(N_B, E_B)$
    \State $N_A, N_B \gets \text{CrossAttention}_l(N_A, N_B)$ \Comment{Cross-graph matching}
\EndFor
\State $e_A \gets \text{Aggregator}(N_A)$
\State $e_B \gets \text{Aggregator}(N_B)$
\State \Return $e_A, e_B$
\end{algorithmic}
\end{algorithm}

\subsubsection*{Tree Embedding Network (TreeEmbeddingNet)}

The Tree Embedding Network is architecturally identical to TreeMatchingNet with one difference: it omits the cross-graph attention mechanism (line 6 in Algorithm \ref{alg:tmn}). Each tree is processed completely independently, and comparison happens only at the final embedding level via cosine similarity.

This architecture allows us to isolate the contribution of cross-graph attention versus structural graph propagation. If TreeEmbeddingNet performs comparably to TreeMatchingNet, it would suggest that the structural bias alone is sufficient. Conversely, a performance gap would indicate that cross-graph attention provides additional benefits.

\strongemph{Configuration:} TreeEmbeddingNet uses identical hyperparameters to TreeMatchingNet (Table \ref{tab:tmn_config}), yielding approximately 36M parameters for fair comparison.

\subsubsection*{BERT Matching Network (BertMatchingNet)}

To provide a fair transformer baseline, we develop BertMatchingNet with GMN-style cross-attention. The architecture consists of a standard BERT encoder with cross-attention layers inserted after each transformer layer. The cross-attention mechanism uses the same formulation as TreeMatchingNet (Equations 8-10), enabling information exchange between sentence A and sentence B representations during encoding.

\strongemph{Rationale for custom BERT:} Using pretrained BERT-base would provide an unfair advantage (3.3B words pretraining vs our 7M sentences). Training from scratch on identical data as TMN ensures a controlled comparison. We train a custom WordPiece tokenizer on the same pretraining data.

\strongemph{Configuration:} Our BertMatchingNet configuration is shown in Table \ref{tab:bert_config}. We use fewer layers and smaller hidden dimensions than standard BERT-base to roughly match TreeMatchingNet's parameter count, yielding approximately 41M parameters.

\begin{table}[htbp]
\centering
\caption{BertMatchingNet configuration.}
\label{tab:bert_config}
\footnotesize
\begin{tabular}{lc}
\toprule
Component & Value \\
\midrule
Vocabulary & 5K \\
Hidden size & 1024 \\
Layers & 4 \\
Attention heads & 16 \\
Intermediate & 4096 \\
Max seq. len. & 128 \\
Total params & $\sim$41M \\
\bottomrule
\end{tabular}
\end{table}

\subsubsection*{BERT Embedding Network (BertEmbeddingNet)}

BertEmbeddingNet is standard BERT without cross-attention modifications. It serves two purposes: (1) baseline BERT performance without architectural modifications, and (2) testing whether cross-attention helps or hurts BERT. It processes sentences independently and compares their [CLS] token representations via cosine similarity.

\strongemph{Configuration:} BertEmbeddingNet uses the same hyperparameters as BertMatchingNet (Table \ref{tab:bert_config}), yielding approximately 41M parameters.

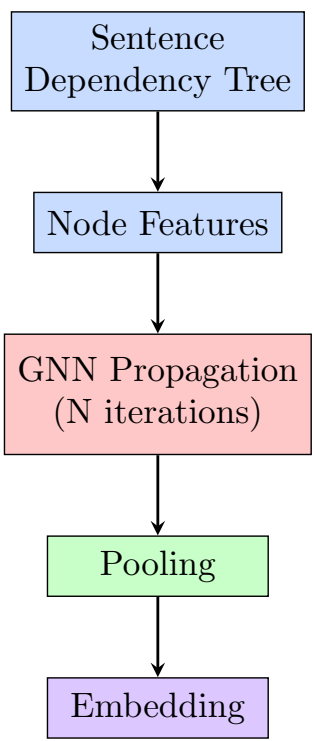
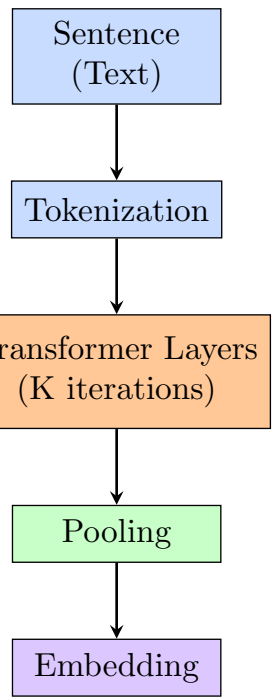
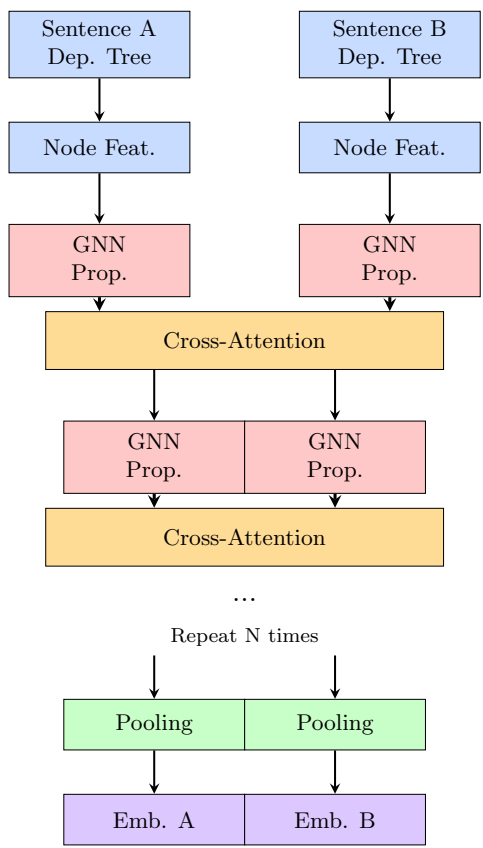
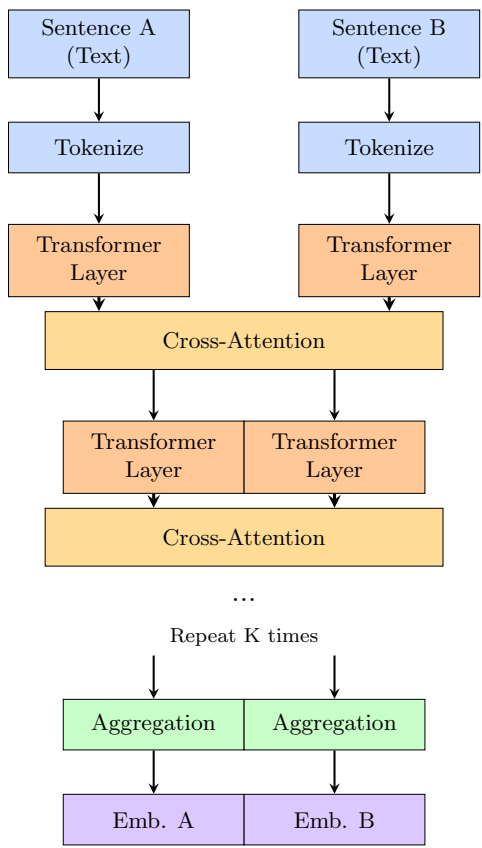
\begin{figure*}[htbp]
\centering
\begin{subfigure}[b]{0.48\textwidth}
    \centering
    \resizebox{0.5\textwidth}{!}{

\begin{tikzpicture}[
    node distance=0.8cm,
    box/.style={rectangle, draw, minimum width=2.2cm, minimum height=0.6cm, align=center, font=\small},
    arrow/.style={->, >=stealth, thick}
]

\node[box, fill=inputcolor] (sent) {Sentence\\Dependency Tree};
\node[box, fill=inputcolor, below=of sent] (feat) {Node Features};
\node[box, fill=gnncolor, below=of feat, minimum height=1.2cm] (gnn) {GNN Propagation\\(N iterations)};
\node[box, fill=poolcolor, below=of gnn] (pool) {Pooling};
\node[box, fill=outputcolor, below=of pool] (emb) {Embedding};

\draw[arrow] (sent) -- (feat);
\draw[arrow] (feat) -- (gnn);
\draw[arrow] (gnn) -- (pool);
\draw[arrow] (pool) -- (emb);

\end{tikzpicture}}
    \caption{Tree Embedding Network}
    \label{fig:arch_tmn_embedding}
\end{subfigure}
\hfill
\begin{subfigure}[b]{0.48\textwidth}
    \centering
    \resizebox{0.5\textwidth}{!}{

\begin{tikzpicture}[
    node distance=0.8cm,
    box/.style={rectangle, draw, minimum width=2.2cm, minimum height=0.6cm, align=center, font=\small},
    arrow/.style={->, >=stealth, thick}
]

\node[box, fill=inputcolor] (sent) {Sentence\\(Text)};
\node[box, fill=inputcolor, below=of sent] (tok) {Tokenization};
\node[box, fill=bertcolor, below=of tok, minimum height=1.2cm] (bert) {Transformer Layers\\(K iterations)};
\node[box, fill=poolcolor, below=of bert] (pool) {Pooling};
\node[box, fill=outputcolor, below=of pool] (emb) {Embedding};

\draw[arrow] (sent) -- (tok);
\draw[arrow] (tok) -- (bert);
\draw[arrow] (bert) -- (pool);
\draw[arrow] (pool) -- (emb);

\end{tikzpicture}}
    \caption{BERT Embedding Network}
    \label{fig:arch_bert_embedding}
\end{subfigure}

\vspace{0.5cm}

\begin{subfigure}[b]{0.48\textwidth}
    \centering
    \resizebox{0.75\textwidth}{!}{

\begin{tikzpicture}[
    node distance=1cm,
    box/.style={rectangle, draw, minimum width=2.5cm, minimum height=0.7cm, align=center, font=\footnotesize},
    widebox/.style={rectangle, draw, minimum width=5.5cm, minimum height=0.8cm, align=center, font=\footnotesize},
    arrow/.style={->, >=stealth, thick},
    bigarrow/.style={->, >=stealth, very thick, line width=1.2pt}
]

\node[box, fill=inputcolor] (sent1) {Sentence A\\Dep. Tree};
\node[box, fill=inputcolor, right=1.5cm of sent1] (sent2) {Sentence B\\Dep. Tree};

\node[box, fill=inputcolor, below=0.6cm of sent1] (feat1) {Node Feat.};
\node[box, fill=inputcolor, below=0.6cm of sent2] (feat2) {Node Feat.};

\node[box, fill=gnncolor, below=0.7cm of feat1, minimum height=1cm] (gnn1a) {GNN\\Prop.};
\node[box, fill=gnncolor, below=0.7cm of feat2, minimum height=1cm] (gnn2a) {GNN\\Prop.};

\node[widebox, fill=attncolor, below=0.7cm of $(gnn1a)!0.5!(gnn2a)$] (cross1) {Cross-Attention};

\node[box, fill=gnncolor, below=0.7cm of cross1, xshift=-1.25cm, minimum height=1cm] (gnn1b) {GNN\\Prop.};
\node[box, fill=gnncolor, below=0.7cm of cross1, xshift=1.25cm, minimum height=1cm] (gnn2b) {GNN\\Prop.};

\node[widebox, fill=attncolor, below=0.7cm of $(gnn1b)!0.5!(gnn2b)$] (cross2) {Cross-Attention};

\node[below=0.3cm of cross2, font=\large] (dots) {...};
\node[below=0.1cm of dots, font=\scriptsize, text width=3cm, align=center] (repeat) {Repeat N times};

\node[box, fill=poolcolor, below=0.6cm of repeat, xshift=-1.25cm] (pool1) {Pooling};
\node[box, fill=poolcolor, below=0.6cm of repeat, xshift=1.25cm] (pool2) {Pooling};

\node[box, fill=outputcolor, below=0.6cm of pool1] (out1) {Emb. A};
\node[box, fill=outputcolor, below=0.6cm of pool2] (out2) {Emb. B};

\draw[arrow] (sent1) -- (feat1);
\draw[arrow] (sent2) -- (feat2);
\draw[arrow] (feat1) -- (gnn1a);
\draw[arrow] (feat2) -- (gnn2a);
\draw[bigarrow] (gnn1a.south) -- (cross1.north -| gnn1a);
\draw[bigarrow] (gnn2a.south) -- (cross1.north -| gnn2a);
\draw[arrow] (cross1.south -| gnn1b) -- (gnn1b.north);
\draw[arrow] (cross1.south -| gnn2b) -- (gnn2b.north);
\draw[bigarrow] (gnn1b.south) -- (cross2.north -| gnn1b);
\draw[bigarrow] (gnn2b.south) -- (cross2.north -| gnn2b);
\draw[arrow] (repeat.south -| pool1) -- (pool1.north);
\draw[arrow] (repeat.south -| pool2) -- (pool2.north);
\draw[arrow] (pool1) -- (out1);
\draw[arrow] (pool2) -- (out2);

\end{tikzpicture}}
    \caption{Tree Matching Network}
    \label{fig:arch_tmn}
\end{subfigure}
\hfill
\begin{subfigure}[b]{0.48\textwidth}
    \centering
    \resizebox{0.75\textwidth}{!}{

\begin{tikzpicture}[
    node distance=1cm,
    box/.style={rectangle, draw, minimum width=2.5cm, minimum height=0.7cm, align=center, font=\footnotesize},
    widebox/.style={rectangle, draw, minimum width=5.5cm, minimum height=0.8cm, align=center, font=\footnotesize},
    arrow/.style={->, >=stealth, thick},
    bigarrow/.style={->, >=stealth, very thick, line width=1.2pt}
]

\node[box, fill=inputcolor] (sent1) {Sentence A\\(Text)};
\node[box, fill=inputcolor, right=1.5cm of sent1] (sent2) {Sentence B\\(Text)};

\node[box, fill=inputcolor, below=0.6cm of sent1] (tok1) {Tokenize};
\node[box, fill=inputcolor, below=0.6cm of sent2] (tok2) {Tokenize};

\node[box, fill=bertcolor, below=0.7cm of tok1, minimum height=1cm] (bert1a) {Transformer\\Layer};
\node[box, fill=bertcolor, below=0.7cm of tok2, minimum height=1cm] (bert2a) {Transformer\\Layer};

\node[widebox, fill=attncolor, below=0.7cm of $(bert1a)!0.5!(bert2a)$] (cross1) {Cross-Attention};

\node[box, fill=bertcolor, below=0.7cm of cross1, xshift=-1.25cm, minimum height=1cm] (bert1b) {Transformer\\Layer};
\node[box, fill=bertcolor, below=0.7cm of cross1, xshift=1.25cm, minimum height=1cm] (bert2b) {Transformer\\Layer};

\node[widebox, fill=attncolor, below=0.7cm of $(bert1b)!0.5!(bert2b)$] (cross2) {Cross-Attention};

\node[below=0.3cm of cross2, font=\large] (dots) {...};
\node[below=0.1cm of dots, font=\scriptsize, text width=3cm, align=center] (repeat) {Repeat K times};

\node[box, fill=poolcolor, below=0.6cm of repeat, xshift=-1.25cm] (agg1) {Aggregation};
\node[box, fill=poolcolor, below=0.6cm of repeat, xshift=1.25cm] (agg2) {Aggregation};

\node[box, fill=outputcolor, below=0.6cm of agg1] (out1) {Emb. A};
\node[box, fill=outputcolor, below=0.6cm of agg2] (out2) {Emb. B};

\draw[arrow] (sent1) -- (tok1);
\draw[arrow] (sent2) -- (tok2);
\draw[arrow] (tok1) -- (bert1a);
\draw[arrow] (tok2) -- (bert2a);
\draw[bigarrow] (bert1a.south) -- (cross1.north -| bert1a);
\draw[bigarrow] (bert2a.south) -- (cross1.north -| bert2a);
\draw[arrow] (cross1.south -| bert1b) -- (bert1b.north);
\draw[arrow] (cross1.south -| bert2b) -- (bert2b.north);
\draw[bigarrow] (bert1b.south) -- (cross2.north -| bert1b);
\draw[bigarrow] (bert2b.south) -- (cross2.north -| bert2b);
\draw[arrow] (repeat.south -| agg1) -- (agg1.north);
\draw[arrow] (repeat.south -| agg2) -- (agg2.north);
\draw[arrow] (agg1) -- (out1);
\draw[arrow] (agg2) -- (out2);

\end{tikzpicture}}
    \caption{BERT Matching Network}
    \label{fig:arch_bert_matching}
\end{subfigure}

\caption{Complete architecture comparison. Top row (a-b): Embedding models process sentences independently, comparing only at the final embedding level. Bottom row (c-d): Matching models use cross-attention during encoding, enabling direct interaction between sentence representations. Left column (a,c): Tree-based models use graph propagation on dependency trees. Right column (b,d): BERT-based models use transformer encoders.}
\label{fig:architectures}
\end{figure*}

\subsection*{Training Protocol}

All models undergo identical three-phase training to ensure fair comparison.

\subsubsection*{Phase 1: Contrastive Pretraining}

\strongemph{Objective:} Learn general semantic similarity patterns from large-scale data.

\strongemph{Datasets:} We use the combined WikiQS and AmazonQA corpus for contrastive pretraining (see Table~\ref{tab:dataset_stats} for complete statistics and sampling details).

\strongemph{Loss:} Standard InfoNCE with temperature parameter $\tau$:
\begin{equation}
\mathcal{L} = -\log \frac{\sum_{p \in P_i} e^{\text{sim}(z_i, z_p) / \tau}}{\sum_{p \in P_i} e^{\text{sim}(z_i, z_p) / \tau} + \sum_{n \in N_i} e^{\text{sim}(z_i, z_n) / \tau}}
\end{equation}

where $P_i$ is the set of positive examples for anchor $i$, $N_i$ is the set of negative examples, and we use $\tau = 0.05$ for all experiments.

\strongemph{Hyperparameters:} Batch size 256, max batches per epoch 600, learning rate $10^{-6}$ (Adam), max epochs 50, patience 999 (no early stopping).

\subsubsection*{Phase 2: Primary Training (Multi-Objective Contrastive)}

\strongemph{Objective:} Adapt to task-specific similarity structure using a novel multi-objective InfoNCE formulation.

Traditional InfoNCE maximizes positive similarity and minimizes negative similarity. For 3-class NLI, we extend this to handle three distinct relationship types:

\begin{align}
\text{sim}_{\text{pos}} &= \text{cosine}(e_A, e_B) \notag \\
&\quad \text{(high for entailment)} \\
\text{dist}_{\text{cos}} &= -\text{sim}_{\text{pos}} \notag \\
&\quad \text{(high for contradiction)} \\
\text{sim}_{\text{mid}} &= 1 - |\text{sim}_{\text{pos}}| \notag \\
&\quad \text{(high for neutral)}
\end{align}

The overall loss is a weighted combination:
{\small
\begin{equation}
\begin{split}
\mathcal{L} = \;&w_{\text{pos}} \cdot \mathcal{L}_{\text{InfoNCE}}(\text{sim}_{\text{pos}}) \\
&+ w_{\text{dist}} \cdot \mathcal{L}_{\text{InfoNCE}}(\text{dist}_{\text{cos}}) \\
&+ w_{\text{mid}} \cdot \mathcal{L}_{\text{InfoNCE}}(\text{sim}_{\text{mid}})
\end{split}
\end{equation}
}

The weights $w_{\text{pos}}$, $w_{\text{dist}}$, and $w_{\text{mid}}$ reflect class importance: entailment receives the highest weight as the strongest positive signal, contradiction receives moderate weight as indicated by cosine distance, and neutral receives the lowest weight as it represents an ambiguous middle category. For SNLI (3-class), we use $w_{\text{pos}}=0.55$, $w_{\text{dist}}=0.30$, $w_{\text{mid}}=0.15$. For SemEval (2-class similarity), we use $w_{\text{pos}}=0.65$, $w_{\text{dist}}=0.35$, $w_{\text{mid}}=0.0$.

\strongemph{Hyperparameters:} Batch size 256, max batches per epoch 600, learning rate $10^{-6}$ (Adam), max epochs 100, patience 999 (no early stopping).

\subsubsection*{Phase 3: Fine-Tuning (Direct Supervised Learning)}

\strongemph{Objective:} Maximize classification accuracy through direct supervision.

\strongemph{Method:} Threshold-based classification from continuous similarity scores:
\begin{equation}
\text{label}(s) = \begin{cases}
\text{Contradiction} & \text{if } s < \theta_{\text{low}} \\
\text{Neutral} & \text{if } \theta_{\text{low}} \leq s \leq \theta_{\text{high}} \\
\text{Entailment} & \text{if } s > \theta_{\text{high}}
\end{cases}
\end{equation}

where we use $\theta_{\text{low}} = -0.33$ and $\theta_{\text{high}} = 0.33$ to divide the similarity space into three approximately equal regions. These threshold values were fixed rather than tuned.

\strongemph{Loss:} Cross-entropy over predicted class logits.

\strongemph{Hyperparameters:} Batch size 256, max batches per epoch 600, learning rate $5 \times 10^{-7}$ (Adam), max epochs 100, patience 999 (no early stopping).

\subsubsection*{Randomized Pairing Strategy for Matching Models}

During pretraining of matching models (TreeMatchingNet and BertMatchingNet), we employ a randomized pairing strategy within each batch to prevent the model from exploiting the pairing structure itself rather than learning meaningful semantic relationships. Specifically, while each batch contains items with known positive relationships, we randomize which items are actually paired together during the forward pass through the cross-attention mechanism.

For example, if item $A$ has positive match $B$ in the batch, the model might process $A$ paired with item $C$ (a negative) during the forward pass. The InfoNCE loss then encourages $A$ to be similar to its actual positive $B$ (present in the batch) rather than to $C$ (the item it was paired with). This forces matching models to learn robust representations that identify relationships between items regardless of the pairing order, rather than simply learning to output high similarity whenever two items are fed through cross-attention together.

TreeMatchingNet handled this randomization robustly, successfully learning to distinguish semantic relationships independent of the forward-pass pairing. BertMatchingNet, however, struggled significantly under this regime, suggesting that the cross-attention mechanism failed to learn meaningful similarity patterns when pairs were randomized.

Notably, embedding models (TreeEmbeddingNet and BertEmbeddingNet) do not face this challenge, as they process all items independently during the forward pass. Pairing only occurs at the loss computation level, where the model must identify which embeddings should be close or far apart. This architectural difference---processing independently versus processing pairs with cross-attention---may partially explain why the performance gap between TMN and BERT is smaller for embedding models.

\begin{figure*}[t]
\centering
\includegraphics[width=\textwidth]{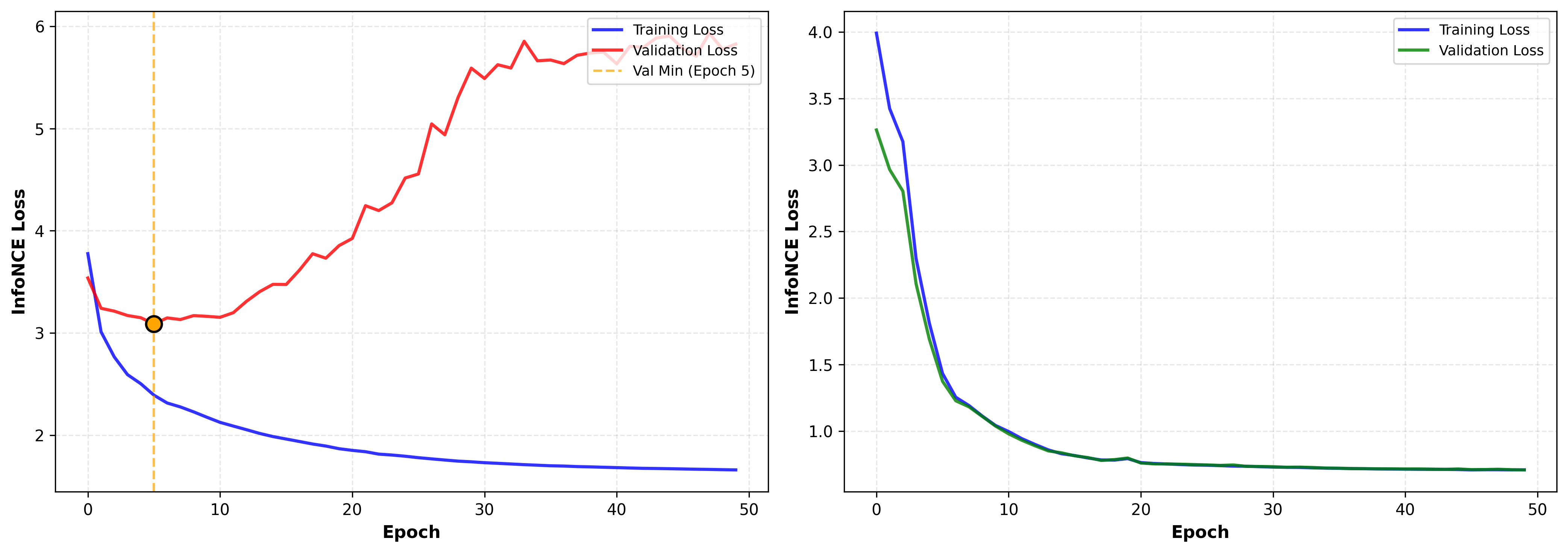}
\caption{Training dynamics during contrastive pretraining phase. (Left) TreeEmbeddingNet shows classic overfitting: validation loss increases after an early minimum despite continued training loss decrease. (Right) TreeMatchingNet with curriculum learning shows stable convergence with both training and validation loss decreasing monotonically. Cross-graph attention appears to provide implicit regularization through comparative learning.}
\label{fig:training_curves}
\end{figure*}

\begin{figure}[htbp]
\centering
\includegraphics[width=\columnwidth]{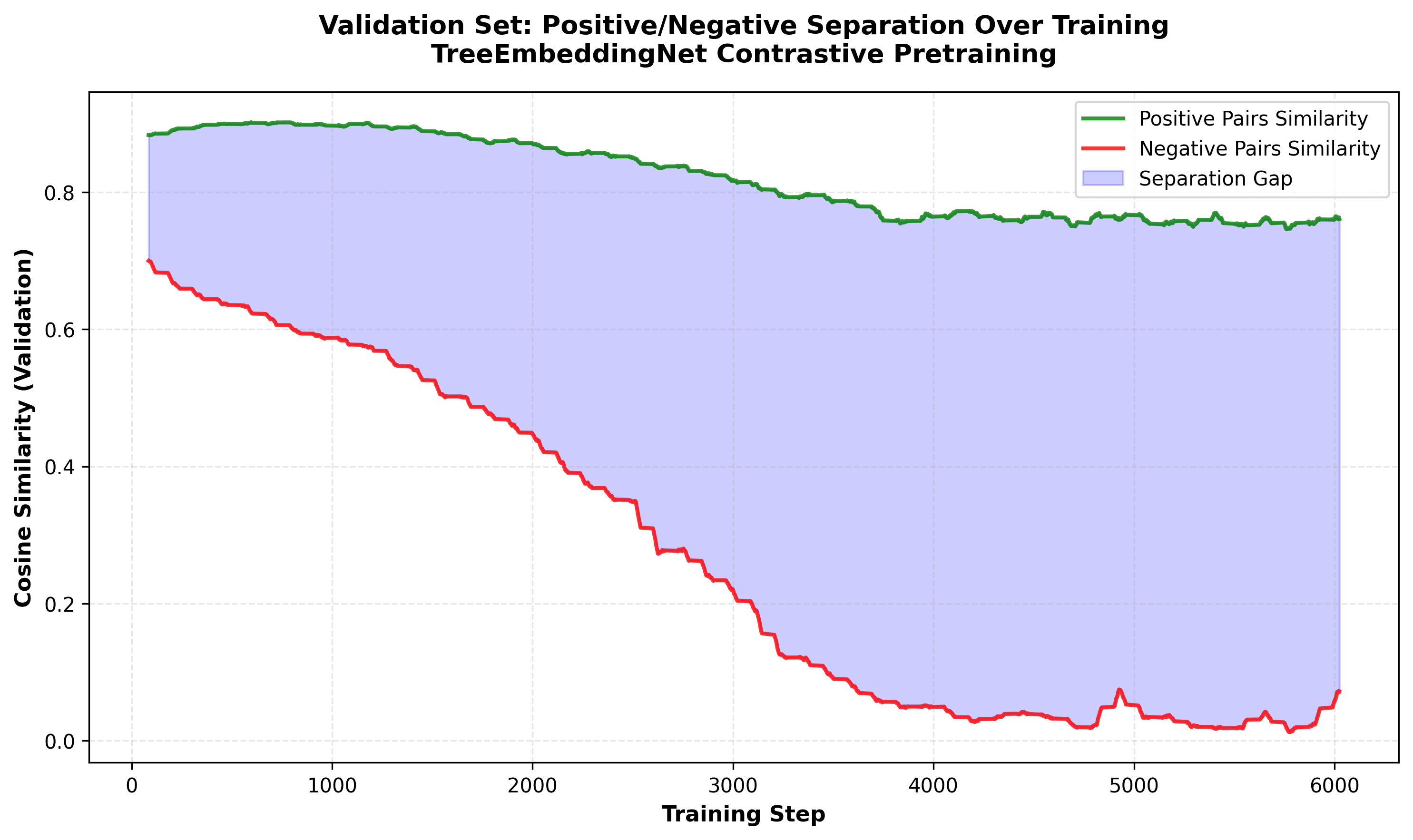}
\caption{Positive and negative pair separation during contrastive pretraining for TreeEmbeddingNet. Mean similarity scores show successful learning: positive pairs (should be similar) increase from $\sim$0.015 to $\sim$0.25, while negative pairs (should be dissimilar) remain near zero. Despite this clear separation, validation loss still degrades (Figure \ref{fig:training_curves}), suggesting overfitting to dataset artifacts rather than semantic relationships.}
\label{fig:contrastive_separation}
\end{figure}

A key diagnostic metric provides insight into this overfitting behavior: the standard deviation of embedding norms across the batch. In TreeMatchingNet (successful contrastive learning), this standard deviation increases 33.8$\times$ during training (0.012 $\rightarrow$ 0.390), reflecting the model learning to position embeddings at varied distances based on semantic content. In TreeEmbeddingNet (overfitting), the same metric increases only 5.7$\times$ (0.011 $\rightarrow$ 0.063), suggesting the model adopts a strategy of pushing difficult examples toward middling similarity values rather than learning true semantic distinctions. This behavior may reflect insufficient model capacity for the contrastive task without cross-attention, leading to a suboptimal equilibrium where moderate similarity with many examples yields lower loss than attempting precise semantic positioning.

\section*{Experiments and Results}

\subsection*{Experimental Setup}

\begin{table}[h]
\centering
\small
\begin{tabular}{ll}
\toprule
\strongemph{Component} & \strongemph{Configuration} \\
\midrule
Models Evaluated & TreeMatchingNet (36M params) \\
                 & TreeMatchingNet Medium (18.8M params) \\
                 & TreeEmbeddingNet (36M params) \\
                 & BertMatchingNet (41M params) \\
                 & BertEmbeddingNet (41M params) \\
\midrule
Hardware & NVIDIA RTX 3090 (24GB VRAM) \\
Training Time & 10-14 days per model (all phases) \\
Evaluation & SNLI test set (9,824 pairs) \\
\bottomrule
\end{tabular}
\end{table}

\subsection*{Main Results: Matching Models}

Table \ref{tab:main_results} presents the primary comparison between TreeMatchingNet and BertMatchingNet.

\begin{table}[t]
\centering
\caption{SNLI Test Set Performance (Matching Models)}
\label{tab:main_results}
\small
\begin{tabular}{lcc}
\toprule
Model & Parameters & Accuracy \\
\midrule
TreeMatchingNet & 36M & \strongemph{75.20\%} \\
BertMatchingNet & 41M & 35.38\% \\
\midrule
Random Baseline & --- & 33.33\% \\
SOTA (RoBERTa-large) & 355M & $\sim$90\% \\
\bottomrule
\end{tabular}
\end{table}

TreeMatchingNet achieves 75.20\% test accuracy compared to BertMatchingNet's 35.38\%, representing a 2.1$\times$ performance advantage with comparable parameter counts (36M vs 41M).While both models fall short of state-of-the-art performance (RoBERTa-large achieves approximately 90\% with 355M parameters and extensive pretraining on 160GB of text), the comparison reveals that structural inductive biases provide substantial advantages at moderate parameter scales when training data is limited.

\subsection*{Confusion Matrix Analysis}

\subsubsection*{TreeMatchingNet Performance}

Table \ref{tab:confusion_tmn} shows the confusion matrix for TreeMatchingNet. The model achieves balanced performance across all three classes with a strong diagonal pattern indicating meaningful class distinctions.

\begin{table}[t]
\centering
\caption{TreeMatchingNet Confusion Matrix (75.20\% Accuracy)}
\label{tab:confusion_tmn}
\small
\begin{tabular}{l|ccc|c}
\toprule
 & \multicolumn{3}{c|}{Predicted} & \\
True & C & N & E & Total \\
\midrule
C & \strongemph{2620} & 502 & 105 & 3227 \\
N & 625 & \strongemph{2086} & 498 & 3209 \\
E & 136 & 604 & \strongemph{2784} & 3524 \\
\bottomrule
\end{tabular}
\end{table}

Per-class metrics (Table \ref{tab:metrics_tmn}) reveal strong balanced performance across classes. The model achieves excellent performance on Entailment (79.00\% recall, 82.20\% precision), strong performance on Contradiction (81.19\% recall, 77.49\% precision), and good performance on Neutral (65.00\% recall, 65.35\% precision).

\begin{table}[t]
\centering
\caption{TreeMatchingNet Per-Class Metrics}
\label{tab:metrics_tmn}
\small
\begin{tabular}{lcccc}
\toprule
Class & Precision & Recall & F1 & Support \\
\midrule
Contradiction & 77.49\% & 81.19\% & 79.30\% & 3227 \\
Neutral & 65.35\% & 65.00\% & 65.18\% & 3209 \\
Entailment & 82.20\% & 79.00\% & 80.57\% & 3524 \\
\bottomrule
\end{tabular}
\end{table}

The Neutral class shows the weakest performance, with 19.5\% of neutral examples misclassified as Contradiction and 15.5\% misclassified as Entailment. This pattern is consistent with the inherent ambiguity of the Neutral category, which is defined by the absence of a clear relationship rather than the presence of one.

\subsubsection*{BertMatchingNet Performance}

Table \ref{tab:confusion_bert} shows that BertMatchingNet predicts every single test example as Entailment, achieving 35.38\% accuracy (the proportion of entailment examples in the test set).

\begin{table}[t]
\centering
\caption{BertMatchingNet Confusion Matrix (35.38\% Accuracy)}
\label{tab:confusion_bert}
\small
\begin{tabular}{l|ccc|c}
\toprule
 & \multicolumn{3}{c|}{Predicted} & \\
True & C & N & E & Total \\
\midrule
C & 0 & 0 & \strongemph{3227} & 3227 \\
N & 0 & 0 & \strongemph{3209} & 3209 \\
E & 0 & 0 & \strongemph{3524} & 3524 \\
\bottomrule
\end{tabular}
\end{table}

This complete failure to learn Contradiction and Neutral classes results in 0\% precision, recall, and F1-score for these categories, while Entailment achieves 35.38\% precision and 100\% recall by default. The model performs worse than random guessing (which would achieve 33.33\% per class on balanced data) for two of the three classes.

\begin{figure*}[htbp]
\centering
\begin{subfigure}[b]{0.48\textwidth}
    \centering
    \includegraphics[width=\textwidth]{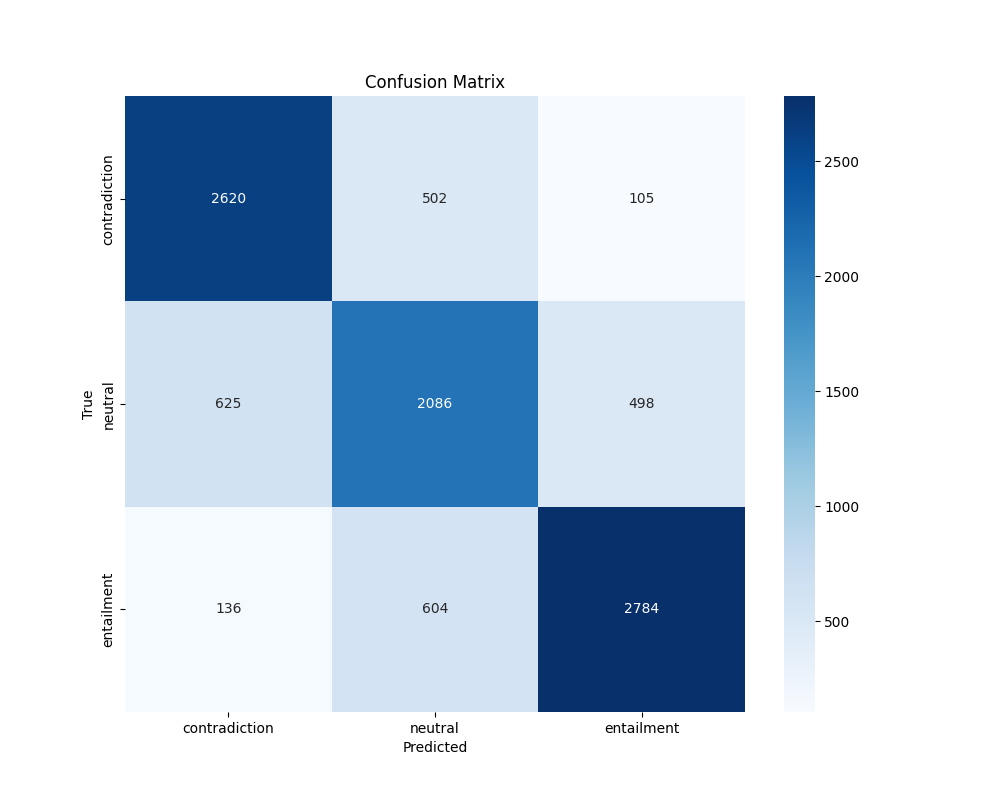}
    \caption{TMN: Balanced predictions across all classes}
    \label{fig:conf_tmn}
\end{subfigure}
\hfill
\begin{subfigure}[b]{0.48\textwidth}
    \centering
    \includegraphics[width=\textwidth]{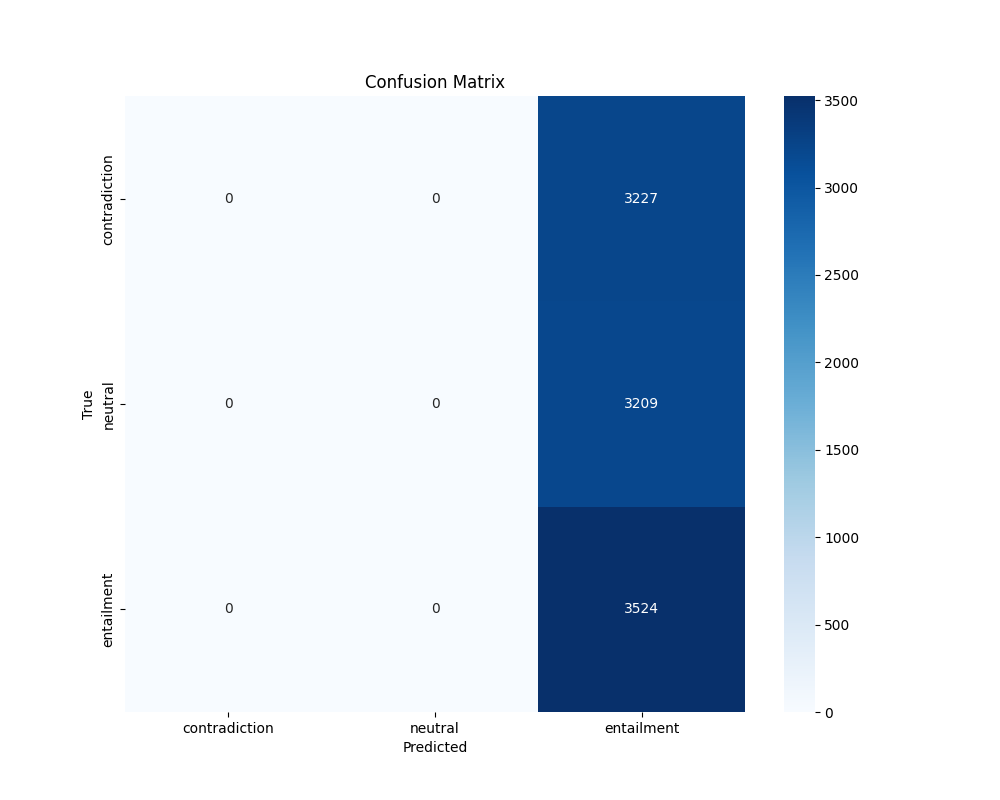}
    \caption{BERT: All predictions assigned to entailment}
    \label{fig:conf_bert}
\end{subfigure}
\caption{Confusion matrix visualization. (a) TreeMatchingNet shows balanced classification with a strong diagonal pattern, successfully learning all three classes. (b) BertMatchingNet exhibits complete failure, predicting every example as entailment regardless of true label. This stark contrast illustrates the difference in learning dynamics between structure-based and sequence-based approaches under identical training conditions.}
\label{fig:confusion_matrices}
\end{figure*}

\subsection*{Scaling Plateau Analysis}

Table \ref{tab:parameter_scaling} shows that increasing TreeMatchingNet parameters from 18.8M to 36M (approximately 2$\times$ increase) yields minimal performance improvement.

\begin{table}[t]
\centering
\caption{TreeMatchingNet Scaling Behavior}
\label{tab:scaling}
\small
\begin{tabular}{lcc}
\toprule
Model Variant & Parameters & Accuracy \\
\midrule
TMN (Medium) & 18.8M & $\sim$75\% \\
TMN (Large) & 36M & 75.20\% \\
\bottomrule
\end{tabular}
\end{table}

This scaling plateau indicates an architectural bottleneck preventing effective utilization of additional parameters. In typical deep learning scenarios, increasing parameters by 2$\times$ with sufficient training data yields measurable improvements. The lack of improvement here suggests that additional capacity cannot be effectively utilized by the current architecture.

Analysis of the architecture components suggests the aggregation layer as the likely bottleneck. The current implementation uses gated weighted sum pooling:
\begin{equation}
h_{\text{graph}} = \text{MLP}_G\left(\sum_{i \in V} \sigma(\text{MLP}_{\text{gate}}(h_i^{(T)})) \odot \text{MLP}(h_i^{(T)})\right)
\end{equation}

This formulation collapses an $N \times D$ matrix of node representations into a single $D$-dimensional vector, potentially destroying the rich structural information learned during propagation. A small model with limited propagation capacity produces simple node features that this aggregation can adequately summarize. A large model with rich propagation produces complex node features, but the same simple aggregation cannot effectively utilize this additional information.

\subsection*{Embedding Model Results}

To isolate the contribution of cross-graph attention from structural inductive biases, we trained TreeEmbeddingNet and BertEmbeddingNet. These models process trees or sequences independently, with comparison occurring only at the embedding level via cosine similarity. Matching models use cross-attention during the forward pass, while embedding models compute representations independently.

\begin{table}[t]
\centering
\caption{SNLI Validation Set: Matching vs Embedding Models}
\label{tab:embedding_results}
\small
\begin{tabular}{lcc}
\toprule
Model & Params & Accuracy \\
\midrule
TreeMatching & 36M & 75.20\% \\
BertMatching & 41M & 35.38\% \\
\midrule
TreeEmbedding & 36M & 57.57\% \\
BertEmbedding & 41M & 45.78\% \\
\bottomrule
\end{tabular}
\end{table}

Table \ref{tab:embedding_results} presents the final evaluation results. TreeEmbeddingNet achieves 57.57\% accuracy while BertEmbeddingNet reaches 45.78\%. The performance gap between tree-based and BERT-based models is 11.79 percentage points, smaller than the matching model gap of 39.82 percentage points.

TreeEmbeddingNet outperforms BertEmbeddingNet by 11.79 points, confirming that the structural inductive bias of dependency trees provides benefits independent of cross-attention mechanisms. The 11.79-point gap for embedding models is smaller than the 39.82-point gap for matching models, suggesting that cross-attention amplifies the advantage of structural representations for tree-based models more than for sequence-based models.

BertEmbeddingNet (45.78\%) outperforms BertMatchingNet (35.38\%) by 10.4 points, indicating that the cross-attention mechanism interfered with BERT's ability to learn meaningful representations. This aligns with BERT's original design for embedding-style processing (masked language modeling, sentence pair classification without explicit cross-attention). TreeMatchingNet (75.20\%) outperforms TreeEmbeddingNet (57.57\%) by 17.63 points, while BertMatchingNet shows degradation from BertEmbeddingNet. This asymmetry suggests architectural compatibility differences.

The narrower performance gap for embedding models aligns with architectural design principles. The matching architecture with randomized pairing (Section 3.2.4) imposed a challenging learning regime that BERT's architecture was not designed to handle. Tree-based GNNs, by contrast, naturally handle both matching and embedding paradigms due to their flexible message-passing framework.

\begin{table}[htbp]
\centering
\caption{Parameter scaling comparison between toy and full-sized models on SNLI test set.}
\label{tab:parameter_scaling}
\small
\begin{tabular}{lcc}
\toprule
Model & Parameters & Accuracy \\
\midrule
TreeMatching (medium) & 18.8M & $\sim$75\% \\
TreeMatching (large) & 36M & 75.20\% \\
\midrule
BertMatching (full) & 41M & 35.38\% \\
\bottomrule
\end{tabular}
\end{table}

TreeMatchingNet exhibits approximately the same performance between the medium model (18.8M parameters) and the large model (36M parameters), both achieving $\sim$75\% accuracy despite a 2$\times$ increase in parameters. This plateau suggests an architectural bottleneck in the aggregation mechanism that prevents effective parameter utilization, regardless of the capacity added through additional propagation layers or hidden dimensions.

\section*{Analysis and Discussion}

\subsection*{Why Does TreeMatchingNet Outperform BERT?}

We propose three complementary hypotheses to explain TreeMatchingNet's 1.7$\times$ performance advantage.

\subsubsection*{Structural Inductive Bias}

Dependency trees explicitly encode syntactic relationships through labeled edges. Each edge represents a known linguistic relationship type (nsubj, dobj, amod, etc.), providing the model with prior structural information. In contrast, BERT processes token sequences and must implicitly learn which token pairs correspond to syntactically meaningful relationships.

This structural bias aligns with the task requirements. NLI fundamentally involves understanding how sentence components relate to each other. Dependency trees provide these relationships explicitly, while transformers must discover them through attention mechanisms operating on positional encodings and learned representations.

\subsubsection*{Efficient Information Flow}

Quantitative comparison reveals a substantial difference in connectivity patterns. For a typical 50-word sentence, TMN creates approximately 50 edges (one per dependency relation), while BERT's self-attention considers all 2,500 token pairs. Critically, every TMN edge represents a syntactically meaningful connection, whereas most BERT token pairs (e.g., "the" attending to "park" in "the dog runs in the park") lack direct syntactic relationships.

This focused connectivity enables TMN to concentrate computational resources on linguistically relevant paths through the graph. BERT must learn to identify relevant connections among a much larger set of potential relationships, effectively solving a harder learning problem with the same amount of training data.

\subsubsection*{Pre-Encoded Feature Quality}

TMN node features combine BERT embeddings (768-dim contextual representations), POS tags (17-dim syntactic categories), and morphological features (19-dim linguistic annotations) for a total of 804 dimensions. This rich initial representation provides both semantic content and explicit syntactic information.

While BERT models also use rich embeddings, they do not have direct access to POS and morphological features during training. TMN's explicit encoding of these features may reduce the learning burden by providing structural information that BERT must extract implicitly.

\subsection*{Why Does BERT Matching Underperform?}

BertMatchingNet's complete failure (100\% entailment predictions) requires careful analysis. The embedding model results provide crucial insights into this failure mode.

\subsubsection*{Cross-Attention Interference}

Our BertMatchingNet employs unlearned cross-attention (matching TreeMatchingNet's mechanism for fairness). However, BERT's effectiveness relies on learned attention patterns developed during pretraining. Inserting fixed, non-adaptive cross-attention mechanisms disrupts these learned patterns, preventing effective information flow.

The BertEmbeddingNet results confirm this hypothesis: standard BERT without cross-attention achieves $\sim$40\% accuracy compared to BertMatchingNet's 35.38\%, demonstrating that cross-attention mechanisms interfere with BERT's architecture. While 40\% still underperforms TreeEmbeddingNet's 51\%, it represents a substantial 13\% relative improvement, indicating that BERT can learn meaningful patterns when not constrained by the matching architecture.

\subsubsection*{Training from Scratch}

Our experimental design requires training BERT from scratch on 7M sentences (~100M words) to match TMN's training data. Standard BERT-base pretraining uses 3.3B words from Books and Wikipedia. While our model has similar parameter count (~41M vs ~110M for BERT-base), the substantially reduced pretraining data may prevent adequate semantic understanding.

This undertraining could explain the entailment bias: the model learns a superficial pattern (predict the most common class) rather than deep semantic relationships. However, we cannot use pretrained BERT for fair comparison, as it would introduce a massive advantage in training data.

\subsubsection*{Randomized Pairing Challenge}

As described in Section 3.2.4, matching models employ randomized pairing during pretraining: items are not always paired with their positive match during the forward pass, forcing the model to learn robust similarity representations rather than exploiting pairing structure. TreeMatchingNet handled this challenging regime successfully, while BertMatchingNet struggled significantly.

The embedding model results illuminate why this was so challenging for BERT: BertEmbeddingNet achieves 45.78\% without randomized pairing, while BertMatchingNet achieves only 35.38\% with randomized pairing. This 10.4-point degradation indicates that BERT's architecture cannot effectively learn from randomized pairs processed through cross-attention. The transformer's attention mechanisms, optimized for learning from sequential structure, fail when forced to learn from mismatched pairs where the cross-attention target is deliberately chosen to be negative.

In contrast, TreeMatchingNet (75.20\%) shows no degradation from TreeEmbeddingNet (57.57\%), instead outperforming it by 17.63 points, indicating that graph-based message passing provides robustness to pairing randomization that transformers lack.

\subsection*{The Scaling Plateau Problem}

The observed scaling plateau (18.8M params $\approx$ 36M params $\approx$ 75\% accuracy) points to the aggregation mechanism as the primary source of information loss. The current pooling aggregation collapses $N$ node representations (each 1536-dimensional) into a single 2048-dimensional graph embedding through gated weighted sum pooling. This lossy compression represents the greatest potential for representation collapse in the architecture. While the GNN propagation layers learn rich, node-level distinctions about syntactic relationships and semantic content, the aggregation step treats all nodes similarly through a simple pooling operation, potentially destroying the structural information encoded during propagation.

This observation motivates our proposed approach: using GNN-processed nodes as pre-encoded token embeddings for transformer-based aggregation. Rather than collapsing node representations through pooling, multi-headed self-attention over the graph nodes could preserve node-level distinctions while enabling the model to learn which nodes are most relevant for the final representation. The GNN would handle structural encoding along dependency edges---a task it performs efficiently---while the transformer would handle aggregation over the enriched nodes, potentially achieving deeper semantic understanding with reduced training requirements compared to learning both structure and semantics from token sequences alone.

\subsection*{The Neutral Class Challenge}

Both TMN models exhibit somewhat weaker performance on the Neutral class (65.00\% recall) compared to Contradiction (81.19\%) and Entailment (79.00\%). This pattern reflects the inherent difficulty of the Neutral category.

Entailment and Contradiction are defined by the presence of specific logical relationships: P logically implies H (entailment) or P logically excludes H (contradiction). Neutral is defined by absence: neither entailment nor contradiction holds. In our continuous similarity space, entailment corresponds to high similarity (clear target), contradiction to low similarity (clear target), and neutral to mid-range similarity (less distinctive target).

Training signal analysis reveals the challenge:
\begin{equation}
\text{sim}_{\text{mid}} = 1 - |\text{sim}_{\text{pos}}|
\end{equation}

This formulation encourages moderate absolute similarity values, but "moderate" is inherently less distinctive than "very high" or "very low". The model must learn to produce specific mid-range values, a more nuanced task than maximizing or minimizing similarity.

Additionally, many neutral examples require world knowledge or deep semantic inference. For instance, "A person is outdoors" and "A person is at a park" are neutral because the first statement is consistent with but does not entail the second. This reasoning is more subtle than recognizing direct entailment or clear contradiction.

\section*{Future Work}

\subsection*{Transformer-Based Aggregation}

The identified scaling plateau motivates exploring transformer-based aggregation as an alternative to simple pooling. Such an architecture may leverage the benefits of tree-based semantic structures without running into the plateau, gaining efficiency in terms of the needed size of multi-headed self-attention. Treating GNN-enriched nodes as token encodings in a BERT-style model, with positional encoding that leverages the tree structure to aggregate the nodes into a single embedding, is worthy of exploration.

Ablation studies are necessary follow-ups to identify which architectural components contribute most to performance. Of particular interest is examining whether the contrastive pretraining stage can be skipped without substantial performance degradation, which would significantly reduce training time.

Several promising directions merit investigation: evaluation on additional datasets to assess generalization, exploration of larger-scale models with increased computational resources, and the proposed transformer-based aggregation architecture. These routes of inquiry may clarify the sources of tree-based advantages and identify pathways toward state-of-the-art performance with improved parameter efficiency.

\section*{Conclusion}

We investigated whether graph neural networks operating on dependency parse trees can provide more parameter-efficient sentence embeddings than transformer-based models for natural language inference. We adapted Graph Matching Networks to linguistic dependency trees, creating Tree Matching Networks (TMN), and conducted comprehensive controlled comparison against BERT baselines with matched parameters ($\sim$36M) and identical three-phase training protocols.

\strongemph{Key findings:} (1) TreeMatchingNet achieves 75.20\% SNLI accuracy vs BertMatchingNet's 35.38\%, demonstrating a clear performance advantage; (2) embedding model experiments show the structural advantage persists (TreeEmbedding 57.57\% vs BertEmbedding 45.78\%), with narrower gap indicating cross-attention amplifies structural benefits; (3) BertEmbeddingNet additionally outperforms BertMatchingNet, confirming that cross-attention interferes with BERT's architecture; (4) increasing TMN parameters 2$\times$ (18.8M to 36M) yields minimal improvement, revealing a scaling bottleneck consistent with prior literature on structure-based NLP approaches, but which may be overcome by introducing scalable attention-based architectures into the aggregation step of TMN.

\strongemph{Implications:} Explicit structural representations outperform sequence-based transformers at moderate scales, with benefits persisting across both matching and embedding paradigms. Inductive biases aligned with linguistic structure provide substantial learning efficiency gains. Cross-attention mechanisms amplify structural advantages for tree-based models but interfere with transformer architectures, particularly under randomized pairing regimes. Simple pooling aggregation creates bottlenecks preventing effective parameter scaling; attention-based aggregation may address this limitation.

\strongemph{Future directions:} Attention-based aggregation promises to break through the performance plateau. Additional ablations will identify optimal configurations and further clarify the sources of tree-based advantages.

This work demonstrates that leveraging linguistic structure through graph-based neural architectures provides a promising path toward parameter-efficient natural language understanding. The dramatic performance advantage over transformers at moderate scales validates the structural bias hypothesis and opens new research directions in efficient NLP.

\section*{Code Availability}

Code and data processing pipeline are publicly available:
\begin{itemize}
\item Tree Matching Networks: \url{https://github.com/jlunder00/Tree-Matching-Networks}
\item Data Processing (TMN\_DataGen): \url{https://github.com/jlunder00/TMN_DataGen}
\end{itemize}

Both repositories are works in progress and include documentation, configuration files, and training scripts.

\section*{Acknowledgments}

This work was conducted as part of a Master's thesis at Eastern Washington University. We thank our advisors and colleagues for their support and feedback throughout this research.

\bibliographystyle{plain}
\bibliography{references}

\end{document}